# DescriptorMedSAM: Language-Image Fusion with Multi-Aspect Text Guidance for Medical Image Segmentation

Wenjie Zhang, MS[1], Liming Luo, MS[2], Mengnan He MS[2], Jiarui Hai, MS[3], Jiancheng Ye, PhD[1,2*]

1 Weill Cornell Medicine, Cornell University, New York, New York, USA

2 Northwestern University, Evanston, USA

3 Johns Hopkins University, Baltimore, Maryland, USA

**Corresponding author:**
Jiancheng Ye, PhD
Northwestern University, Evanston, USA
jiancheng.ye@u.northwestern.edu

## ABSTRACT

Accurate organ segmentation is essential for clinical tasks such as radiotherapy planning and disease monitoring. Recent foundation models like MedSAM achieve strong results using point or bounding-box prompts but still require manual interaction. We propose DescriptorMedSAM, a lightweight extension of MedSAM that incorporates structured text prompts, ranging from simple organ names to combined shape and location descriptors to enable click-free segmentation. DescriptorMedSAM employs a CLIP text encoder to convert radiology-style descriptors into dense embeddings, which are fused with visual tokens via a cross-attention block and a multi-scale feature extractor. We designed four descriptor types: Name (N), Name + Shape (NS), Name + Location (NL), and Name + Shape + Location (NSL), and evaluated them on the FLARE 2022 dataset under zero-shot and few-shot settings, where organs unseen during training must be segmented with minimal additional data. NSL prompts achieved the highest performance, with a Dice score of 0.9405 under full supervision, a 76.31% zero-shot retention ratio, and a 97.02% retention ratio after fine-tuning with only 50 labeled slices per unseen organ. Adding shape and location cues consistently improved segmentation accuracy, especially for small or morphologically complex structures. We demonstrate that structured language prompts can effectively replace spatial interactions, delivering strong zero-shot performance and rapid few-shot adaptation. By quantifying the role of descriptor, this work lays the groundwork for scalable, prompt-aware segmentation models that generalize across diverse anatomical targets with minimal annotation effort.

## INTRODUCTION

Accurate segmentation of abdominal organs is critical for tasks such as radiotherapy planning and disease monitoring. Traditional deep learning models, including U-Net[1] and its derivatives (e.g., V-Net[2], PSPNet[3], and TransUNet[4]), achieve high Dice scores when trained on thousands of annotated organ masks. However, generating these annotations is both costly and time-consuming for clinicians[5]. To reduce this burden, researchers have explored interactive segmentation methods that require only a few user-provided prompts[6, 7]. The Segment Anything Model (SAM)[8] enables high-quality segmentation with minimal input, and MedSAM extends this approach to computed tomography (CT) and magnetic resonance imaging (MRI). Yet, even these spatial prompts, such as clicks on small nodules still demand expert involvement. This limitation has motivated the exploration of even more accessible prompts, including natural language descriptions[9].

Large vision–language models, such as Contrastive Language-Image Pre-training (CLIP)[10], bridge visual and textual domains by aligning a Vision Transformer (ViT) image encoder with a text encoder trained on millions of image–caption pairs.[11] This alignment allows free-form text to guide segmentation without explicit clicks. Recent work has applied similar approaches to medical imaging: FLanS[12] generates free-form text to guide segmentation, and STPNet [13] integrates scale-aware text prompts to enhance performance. However, these studies rarely examine how the granularity of textual prompts influences segmentation quality - a key factor for minimizing annotation effort and ensuring reliable clinical use.[14] Moreover, most prior work evaluates only on organs present in training data, leaving the challenges of zero-shot and few-shot generalization to unseen anatomy largely unaddressed.

Building on these observations, we propose DescriptorMedSAM, a lightweight text-guided extension of MedSAM. Our method employs a CLIP text encoder that converts radiology-style descriptors -ranging from simple organ names to detailed prompts such as “Isolate the liver, a large wedge-shaped organ in the right upper quadrant beneath the diaphragm” - into dense embeddings. We systematically study prompt granularity by designing four types of descriptors: Name (N), Name + Shape (NS), Name + Location (NL), and Name + Shape + Location (NSL).

These structured prompts are evaluated under zero-shot and few-shot scenarios, where the model must segment organs unseen during training[15]. In the few-shot setting, fine-tuning with only 50 labeled slices per organ significantly boosts performance, highlighting the potential of structured language to enable rapid adaptation to new anatomical targets. Our contributions are as follows:

1. We introduce a parameter-efficient architecture that integrates linguistic descriptors into MedSAM via a cross-attention mechanism, improving abdominal organ delineation.

2. We develop and evaluate a structured taxonomy of four radiology-style prompts (N, NS, NL, NSL) across 12 abdominal organs from the FLARE-22 dataset[16] under both zero-shot and few-shot settings.

## METHODS

### Dataset

All experiments were conducted on the FLARE 2022 dataset[16], a benchmark comprising 50 contrast‑enhanced abdominal CT volumes with expert annotations for 13 organs: liver, spleen, pancreas, stomach, gallbladder, duodenum, esophagus, aorta, inferior vena cava (IVC), left and right kidneys, and left and right adrenal glands. Each volume is provided in 3D NIfTI format. To adapt this data to our 2D segmentation pipeline, we resampled scans to a uniform pixel spacing and stored them as compressed NumPy archives (.npz), resulting in 24,234 fully labeled slices. For consistency across organs, any class with fewer than 2,000 labeled slices was excluded; consequently, the duodenum was omitted, leaving 12 organs for training, validation, and testing.

### Baseline MedSAM

MedSAM processes 2D slices using a frozen ViT‑B/16 image encoder pretrained on large-scale natural and medical image datasets. The encoder converts input slices into patch‑level visual tokens. A separate prompt encoder embeds user-provided guidance (e.g., points or bounding boxes) into tokens. These visual and prompt tokens are concatenated and passed to a mask decoder, which outputs a probability map for the target structure.

### DescriptorMedSAM Architecture

The full architecture of DescriptorMedSAM is illustrated in Figure 1. It retains MedSAM's frozen image encoder and mask decoder while introducing three lightweight modules: (i) Multi-scale feature extractor, (ii) Cross-attention block, and (iii) CLIP-based text-prompt encoder.

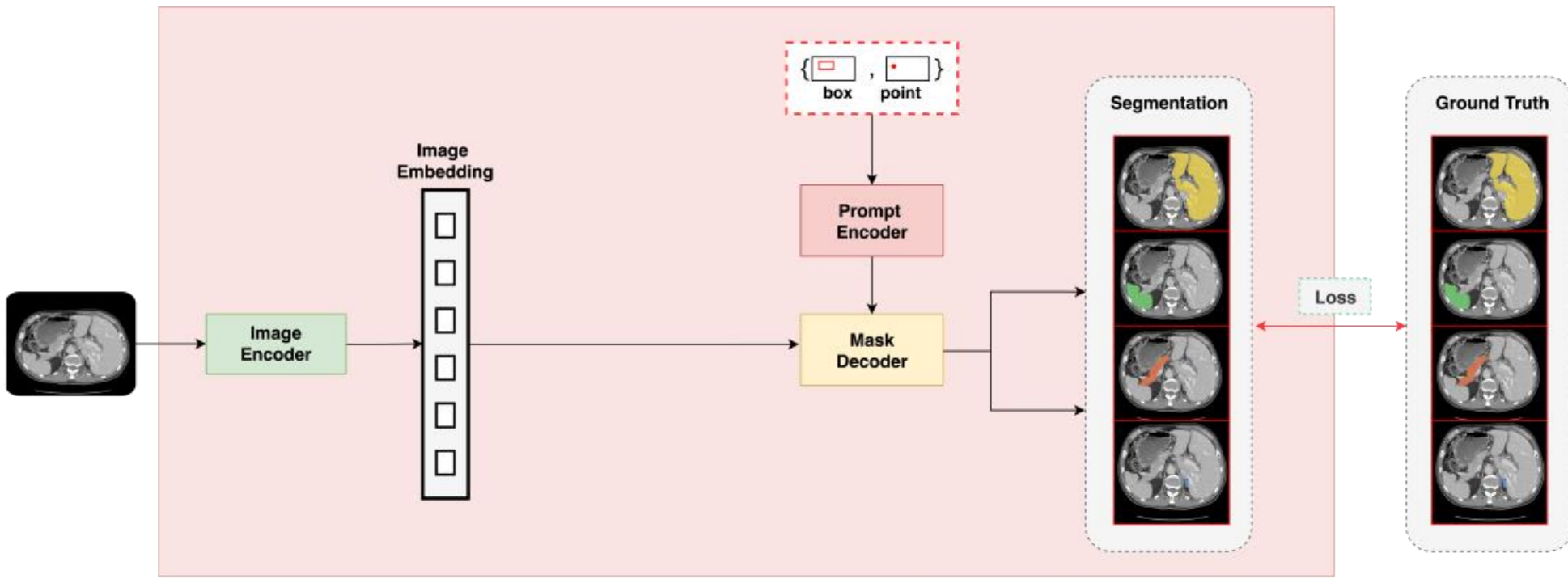

(a) MedSAM

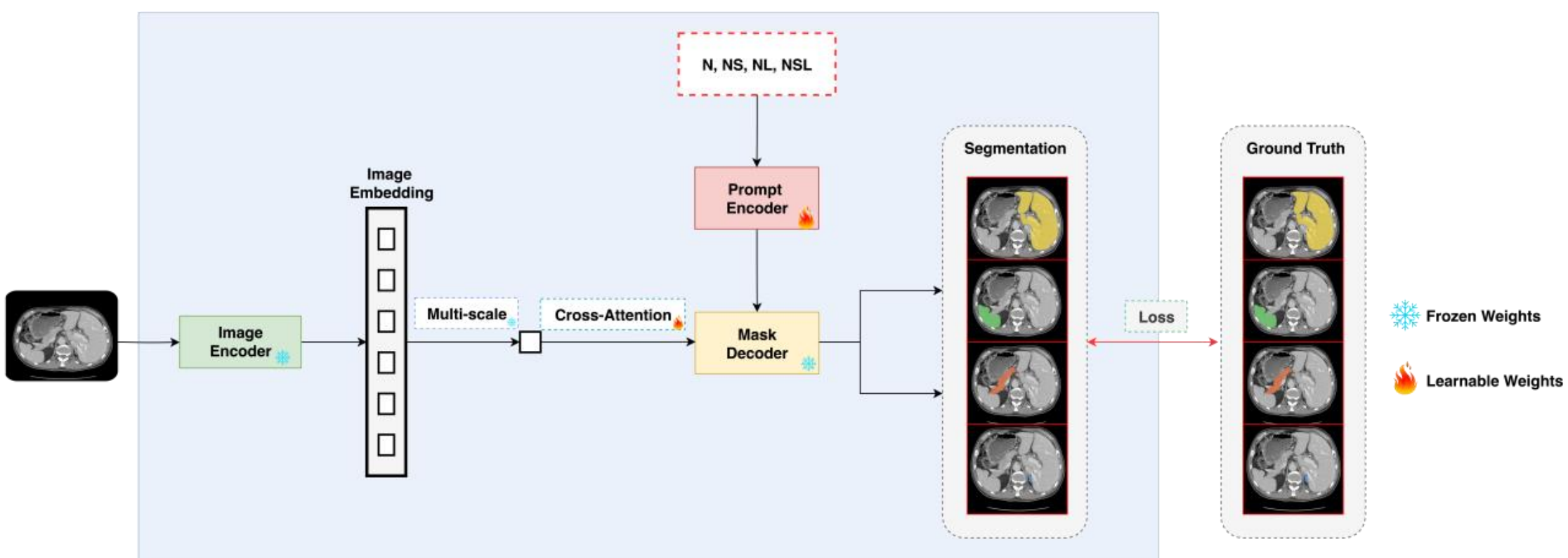

(b) DescriptorMedSAM (ours)

**Figure. 1** Comparison between (a) MedSAM and (b) DescriptorMedSAM.

First, intermediate features from four layers $\{l_1, l_2, l_3\}$ of the ViT backbone[12] are aggregated and concatenated. A 1×1 convolutional neck followed by pixel-shuffling produces a multi-scale feature pyramid, preserving contextual cues while recovering fine spatial details.

Second, radiology-style descriptors are embedded using a CLIP text encoder[10], and the resulting textual tokens are projected to match the dimensionality of visual tokens.

Third, Text embeddings are injected into the visual stream via a multi-head cross-attention mechanism[17]. Let $V \in \mathbb{R}^{N \times d}$ be the visual tokens and $t \in \mathbb{R}^{d_t}$ be the CLIP text embedding. After a linear projection $P_t \in \mathbb{R}^{d_t \times d}$, the block computes

$$
\begin{aligned}
Q &= VW_Q, \\
K &= (P_t t)W_K, \\
U &= (P_t t)W_V, \\
\tilde{V} &= LN(V + MHA(Q, K, U))
\end{aligned}
\tag{1}
$$

Where $W_Q$, $W_K$, $W_V \in \mathbb{R}^{d \times d}$ are learned linear maps, $MHA(\cdot)$ is standard multi-head attention, and $LN(\cdot)$ denotes layer normalization with a residual connection. The enhanced token $\tilde{V}$ are concatenated with the multi-scale features and fed into the mask decoder. This modification adds only 1.9 million trainable parameters (≈1.7% of total model size), maintaining computational efficiency.

We optimize the decoder with a compound loss that sums dice [18]and binary cross-entropy (BCE) :

$$\mathcal{L} = \lambda_D \mathcal{L}_{Dice} + \lambda_B \mathcal{L}_{BCE} \tag{2}$$

Where (with logits $z$ and ground truth mask $y \in \{0,1\}^{H \times W}, \hat{y} = \sigma$(z)):

$$\mathcal{L}_{Dice} = 1 - \frac{2 \sum \hat{y} y + \varepsilon}{\sum \hat{y} + \sum y + \varepsilon} \tag{3}$$

$$\mathcal{L}_{BCE} = -\frac{1}{HW} \sum [y \log \hat{y} + (1 - y) \log(1 - \hat{y})] \tag{4}$$

We set $\lambda_D = \lambda_B = 1$ and $\varepsilon = 10^{-6}$ for numerical stability.

### Radiological Prompt Construction

To translate anatomical knowledge into textual prompts, we designed four descriptor prompt categories: N, NS, NL, and NSL, which progressively incorporate name, shape, and location cues. The examples of prompts are as follows:

**Table 1.** Radiological prompt categories, illustrative examples, and GPT-4 generation templates.

| Prompt Category | Generation prompt to GPT-4 | Example |
|---|---|---|
| N | / | Liver |
| NS | Generate one concise radiology-style sentence that describes the typical shape of the {organ}. | Segment the curved, oval spleen hugging the stomach fundus. |
| NL | Generate one concise sentence that states the anatomical location of the {organ} using neighboring structures as landmarks. | Mark pancreas body crossing anterior to the aorta. |
| NSL | Generate one concise sentence that combines both shape and location information for the {organ}. | Segment the right adrenal gland; note its triangular outline positioned atop the right kidney, posterior to the liver. |

The N prompt contains only the organ's lowercase name, representing the most minimal user query. NS prompts enrich this baseline with morphological cues; a spleen slice might be guided by "Segment the curved, oval spleen hugging the stomach fundus." NL prompts anchor the structure within axial anatomy by citing neighboring organs and directional landmarks, such as "Mark pancreas body crossing anterior to aorta." NSL prompts combine these two aspects into one fluent instruction. For example," segment the right adrenal gland; note its triangular outline positioned atop the right kidney, posterior to the liver."

To generate such sentences, we supply GPT-4[19] with one template per prompt category, substituting the placeholder {organ} with each target-organ name. For every organ, we request 20 variants, resulting in 240 prompts per category (12 organs × 20) and a total of 960 prompts across the four categories. Batched API calls are executed with temperature 0.7, top-p 0.9, and a fixed random seed, guaranteeing that identical calls regenerate the corpus byte-for-byte. The resulting sentences are concise (median ≈ 12 tokens) but provide expressive guidance for every organ.

**Experimental Protocol**

To evaluate the effectiveness of prompt-guided segmentation, we follow a two-stage learning protocol: zero-shot and few-shot learning. In the zero-shot stage, the model is trained using labeled slices from a randomly selected subset of 8 seen organs and evaluated on the 4 remaining unseen organs.

In the Round r, we use $Dice_{ZS}$ to denote the dice on those unseen organs and $Dice_F$ to denote the fully supervised dice. We report the zero-shot retain ratio as

$$RR_{ZS} = \frac{Dice_{ZS}}{Dice_F} \quad (5)$$

In the few-shot stage, following the protocol in SAM Few-shot Finetuning for Anatomical Segmentation in Medical Images[20], we fine-tune the model starting from the zero-shot checkpoint using 50 labeled axial slices per unseen organ. The model is then tested on the remaining slices for those organs. The dice score is $Dice_{FS50}$ , and the few-shot retain ratio is computed as

$$RR_{FS} = \frac{Dice_{FS50}}{Dice_F} \tag{6}$$

To ensure robustness and mitigate sampling bias, we repeat the experiment five times (Rounds 1–5) for every prompt type[21]. The same seed is applied across all prompt types—N, NS, NL, and NSL—ensuring that each strategy is evaluated under identical training and test partitions. Within each seen organ, slices are randomly divided into 80% training and 20% validation sets. All hyperparameters are kept constant across all rounds and prompt variants.

### Implementation

The proposed framework is implemented on the PyTorch library with one NVIDIA 80G H100 GPU. We adopted the original MedSAM as the base and loaded its official weights. During the training and testing step, the model is trained for 30 epochs on the Flare 2022 dataset with a batch size of 8[16], using the Adam optimizer (learning rate: 0.0001). The baseline models are also trained for 30 epochs on the Flare 2022 dataset with the same hyperparameters.

## RESULTS

### Fully-Supervised Evaluation

We compared four text-prompt strategies—N, NS, NL, and NSL—against two strong baselines, SAM[8] and MedSAM[22] across 12 abdominal organs. Dice scores are summarized in Table 2.

**Table 2.** Dice scores per organ for SAM, MedSAM, and DescriptorMedSAM with four prompt types.

| Organ | SAM | MedSAM | DescriptorMedSAM-N | DescriptorMedSAM-NS | DescriptorMedSAM-NL | DescriptorMedSAM-NSL |
|---|---|---|---|---|---|---|
| Liver | 0.9222 | 0.949 | 0.9699 | 0.9819 | 0.9716 | **0.9827** |
| Right kidney | 0.96 | 0.9449 | 0.9648 | **0.9781** | 0.9754 | 0.9764 |
| Spleen | 0.9567 | 0.957 | 0.9662 | **0.9812** | 0.968 | 0.973 |
| Pancreas | 0.8009 | 0.8567 | 0.8629 | 0.8649 | 0.8749 | **0.8903** |
| Aorta | 0.9534 | 0.9397 | 0.9557 | 0.9652 | 0.9644 | **0.9677** |
| Ivc | 0.9272 | 0.9317 | **0.9511** | 0.9454 | 0.947 | 0.9508 |
| Right adrenal gland | 0.7405 | **0.8964** | 0.8672 | 0.8701 | 0.8733 | 0.8755 |
| Left adrenal gland | 0.8084 | **0.8851** | 0.871 | 0.8677 | 0.8717 | 0.8706 |
| Gallbladder | 0.9092 | 0.905 | 0.9257 | **0.9405** | 0.9364 | 0.9378 |
| Esophagus | 0.8601 | **0.9264** | 0.9066 | 0.9158 | 0.9043 | 0.9137 |
| Stomach | 0.8643 | 0.9193 | **0.9713** | 0.9685 | 0.9673 | 0.9695 |
| Left kidney | 0.9539 | 0.9477 | 0.9569 | 0.9773 | 0.9761 | **0.9784** |
| Average Dice | 0.8881 | 0.9216 | 0.9308 | 0.9381 | 0.9359 | **0.9405** |

NSL prompts attained the highest overall performance, achieving an average Dice score of 0.9405 and exceeding the performance of MedSAM and SAM by 1.9 and 5.2 percentage points, respectively. NSL prompts were the top performer for four organs (liver, pancreas, aorta, and left kidney). NS prompts ranked second overall and led in right kidney, spleen, and gallbladder segmentation, indicating that shape information alone significantly enhances delineation. N prompts performed best on IVC and stomach, whereas NL prompts did not lead in any single organ but consistently delivered competitive results. Interestingly, MedSAM outperformed DescriptorMedSAM on three small, low-contrast organs (left adrenal gland, right adrenal gland, and esophagus), suggesting that such structures remain more sensitive to noise in textual prompts. Overall, NSL prompts emerged as the most reliable choice, improving segmentation accuracy across all organs without degrading performance for any specific case.

**Zero- & Few-Shot Adaptation**

**Table 3** summarizes the mean zero-shot and few-shot retention ratios across five experimental rounds.

**Table 3.** Mean zero-shot and 50-shot retain ratios (five-split average). 95 % CIs were derived with a two-tailed t-distribution; upper bounds > 100 % or lower bounds < 0 % were truncated to 100 % and 0 %, respectively.

| Model Type | Zero-shot Retain Ratio | Few-shot Retain Ratio |
|---|---|---|
| DescriptorMedSAM-N | 69.83 % (95 % CI 59.71%–79.96%) | 90.66 % (95 % CI 79.26%–100.00%) |
| DescriptorMedSAM-NS | 71.17 % (95 % CI 60.62%–81.72%) | 93.94 % (95 % CI 86.85%–100.00%) |
| DescriptorMedSAM-NL | 72.51 % (95 % CI 63.99%–81.02%) | 93.39 % (95 % CI 85.62%–100.00%) |
| DescriptorMedSAM-NSL | **76.31 % (95 % CI 63.72%–88.90%)** | **97.02 % (95 % CI 94.20%–99.83%)** |

Adding textual detail consistently improved zero-shot generalization. Compared to N prompts, shape prompts (NS) improved retention by 1.3 percentage points, location prompts (NL) by 2.6 points, and combined prompts (NSL) by 6.5 points. Without additional labels, NSL prompts already preserved 76.31 % of fully supervised Dice scores, narrowing the performance gap to less than 25 %.

With few-shot fine-tuning using only 50 labeled slices per unseen organ, all prompt types achieved substantial gains. NSL prompts reached 97.02 % of fully supervised performance, outperforming NS and NL prompts (~94 %) and N prompts (~91 %). These findings indicate that morphological and spatial cues are complementary, with NSL prompts enabling the fastest and highest adaptation to unseen organs.

Confidence intervals for all prompt types remained within ±15 percentage points, indicating stable performance across experimental rounds. Although some intervals overlap, the NSL prompt distribution consistently lies higher, reinforcing its superior generalization in both zero-shot and few-shot scenarios.

**Figure 2** provides a qualitative comparison of DescriptorMedSAM's outputs under different training regimes. The figure shows that the fully supervised model closely replicates the ground-truth masks across all organs. In the zero-shot setting, DescriptorMedSAM with NSL prompts already covers some of the large, high-contrast organs, such as the stomach and left kidney; however, it under-segments smaller or morphologically complex structures, including the pancreas and both adrenal glands. After fine-tuning with just 50 labelled slices per unseen organ, the red contours for these challenging organs tighten markedly, and the visual results approach those of the fully supervised model. This qualitative improvement confirms that a small annotation budget can substantially narrow the performance gap to full supervision.

Overall, NSL provides the strongest zero-shot safety net and the steepest few-shot learning curve once a handful of labelled slices becomes available.

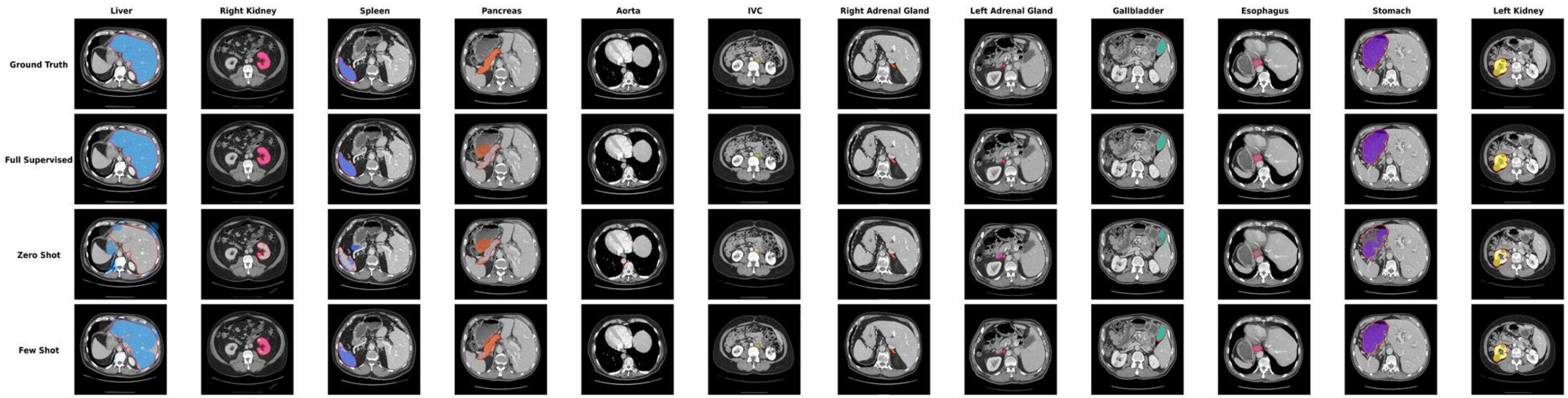

**Figure 2.** Qualitative segmentation results for twelve abdominal organs. Columns list the organs (Liver to Left Kidney); rows show, top to bottom: Ground Truth masks, Fully-Supervised DescriptorMedSAM (NSL), Zero-Shot DescriptorMedSAM (NSL), and Few-Shot DescriptorMedSAM after 50 labelled slices per unseen organ. Ground-truth regions are filled with organ-specific colors; predicted boundaries are overlaid in red.

## DISCUSSION

Recent advances in medical image segmentation have largely centered on interactive spatial prompting, where models such as SAM[8] and MedSAM[22] rely on user‐supplied points or bounding boxes to localize target regions. Variants like ProtoSAM[23] further streamline this interaction. Although these approaches—built on a shared encoder - decoder architecture, delivering state-of-the-art accuracy on organs encountered during training, they share a critical limitation: every new case still demands additional human input. This extra annotation workload hinders scalability and complicates deployment in time-sensitive clinical settings. To break this dependency on clicks, researchers have begun exploring language-based prompting to guide the segmentation. FLanS[12] employs large language models to generate free-form descriptions, STPNet[13] introduces scale-aware text prompts for lesion segmentation, and TGAM fuses anatomical sentences with visual features[24]. These studies demonstrate that natural-language guidance can effectively replace spatial prompts and integrate seamlessly with radiology reporting. However, they share several limitations: most treat the text prompt as a black-box input without examining its semantic granularity, and they primarily evaluate on anatomies already encountered during training.

To overcome these problems, we introduce DescriptorMedSAM—a descriptor-guided extension of MedSAM. It maintains MedSAM's image encoder and mask decoder in a frozen state to preserve speed and memory, while injecting language via a parameter-To unpack the role of linguistic detail, we devise a four-level prompt taxonomy—N, NS, NL, and NSL—that allows us to measure how each semantic aspect influences performance. Finally, we established a unified zero-shot and few-shot protocol on FLARE 2022, utilizing five random organ splits, which enabled rigorous tests on anatomy never encountered during training and rapid adaptation with just 50 labelled slices per unseen organ. In summary, DescriptorMedSAM captures fine-grained textual information, generalizes effectively to unseen anatomy, and maintains the efficiency required for real-time clinical deployment.

Empirically, finer semantics consistently help: NSL improves the zero-shot retain ratio by 6.5 percentage points over N and reaches 97.02% of fully supervised performance after 50 slices per organ of fine-tuning, with results stable across splits (95% CIs within ±15 pp). Qualitative results echo these trends: zero-shot predictions already cover large organs well but miss small or intricate structures, whereas few-shot fine-tuning sharpens the boundaries of those challenging regions, bringing them close to fully supervised quality. These findings clarify that DescriptorMedSAM retains the click-free convenience of language-guided segmentation, while quantifying the impact of prompt semantic granularity and delivering substantial zero-shot and few-shot improvements on previously unseen organs.

By systematically quantifying the effect of semantic granularity, DescriptorMedSAM is the first to show that combining shape and location cues provides distinctive benefits for delineating organ boundaries. At the same time, it delivers strong zero-shot performance from structured text alone and rapidly approaches fully supervised accuracy with just a handful of labeled slices. This capability can shorten model-development cycles from weeks to days and substantially ease the burden on clinicians.

## Limitation

This study has several limitations. First, DescriptorMedSAM was evaluated on a relatively small dataset of 50 CT volumes, raising questions about generalizability to more heterogeneous populations spanning multiple scanners, institutions, and pathologies. Future work should validate the approach on larger, multi-institutional datasets and across different imaging modalities.[25] Second, our prompt design focused on four fixed categories (N, NS, NL, NSL), which may not capture nuances such as pathology-specific descriptors, temporal context, or radiologist shorthand. Expanding to adaptive or pathology-aware prompts could further enhance performance and robustness. Finally, while our architecture is lightweight, real-time clinical deployment will require additional work to optimize inference speed, integrate into Picture Archiving and Communication System (PACS), and evaluate usability with radiologists in prospective workflows.

## CONCLUSION

This study presents DescriptorMedSAM, an extension of MedSAM that integrates structured descriptor prompts into medical image segmentation via a cross-attention mechanism and multi-scale feature fusion. By systematically varying prompt granularity—from simple organ names to combined shape-location descriptors—we demonstrate that richer textual guidance significantly enhances both zero-shot and few-shot segmentation performance. By quantifying the impact of prompt semantics and demonstrating strong generalization to unseen organs, DescriptorMedSAM establishes a foundation for prompt-aware medical segmentation models. Future work should extend this framework to larger, multi-institutional datasets, additional imaging modalities, and adaptive prompt generation strategies, further advancing the clinical applicability of language-guided segmentation.

## Conflict of Interest Statement

None.

## Contribution Statement

JY designed the study. WZ, JH, LL, and JY contributed to data analyses. WZ and JY contributed to the writing of the manuscript. MH contributed to data management. All authors read and approved the final version of the manuscript.

## Data Availability Statement

FLARE22 are available at https://flare22.grand-challenge.org.

## Code Availability Statement

The relevant code are available at: https://github.com/Wenj1eee/Prompt-Dimensions-of-MedSAM

## Funding

None.